\documentclass[11pt]{article}
\usepackage[preprint]{acl}
\usepackage{times}
\usepackage{latexsym}
\usepackage[T1]{fontenc}
\usepackage[utf8]{inputenc}
\usepackage{microtype}
\usepackage{inconsolata}
\usepackage{amsmath}
\usepackage{booktabs}
\usepackage{array}
\usepackage{graphicx}
\hypersetup{pdftitle={Nothing Left Behind: Embedding Retrieval and Language Model Screening for High Recall Evidence Search},
  pdfauthor={Development Impact AI Lab, World Bank Group}}

\title{Nothing Left Behind: Embedding Retrieval and Language Model\\Screening for High Recall Evidence Search}

\author{
 \textbf{Giuliano Martinelli}\textsuperscript{*},
 \textbf{Abelardo Carlos Martinez Lorenzo},
 \textbf{Satvik Garg},
 \textbf{Linxi Wang},
\\
 \textbf{Riccardo Orlando},
 \textbf{Piriyakorn Piriyatamwong},
 \textbf{Samuel Paul Fraiberger}\textsuperscript{\dag}
\\
 Development Impact AI Lab, World Bank Group
\\
 \small{\textsuperscript{*}Lead author. \textsuperscript{\dag}Correspondence: \href{mailto:sfraiberger@worldbankgroup.org}{sfraiberger@worldbankgroup.org}}
}

\begin{document}
\maketitle

\begin{abstract}
AI systems that ground their answers from retrieved documents usually need only a few relevant passages, but applications such as systematic reviews and legal discovery need nearly all of them, because their conclusions rest on the full body of evidence. 
These applications have traditionally searched with curated taxonomies, which can only find what their vocabulary describes. 
We compare taxonomy-based retrieval with an alternative neural retrieval based on sentence embeddings and followed by language model screening in ImpactAI, an evidence synthesis application built on 7,354 randomized trials. 
Measured against the studies included in 29 systematic reviews, the share of studies found rises from 45\% to 85\%, and the screening model agrees with human annotators as well as annotators agree with each other. 
This gain comes at a cost in speed: at the median, retrieving and screening evidence for a query takes 77 seconds, against 6 seconds for taxonomy retrieval. 
The results suggest finding evidence with embeddings, organizing it with taxonomy labels, and setting how many candidates to screen according to the time available for each query.
\end{abstract}

\section{Introduction}
\label{sec:introduction}

Language models that answer from retrieved documents, the setup known as retrieval augmented generation (RAG)~\citep{lewis2020rag}, usually need only a few relevant passages to answer a question. Some applications, however, need every relevant document, because their output is a conclusion about the whole body of evidence. A systematic review, for example, pools all trials of an intervention to estimate its effect, so a missed trial can shift both the estimate and the conclusion drawn from it. Similarly, in legal discovery a party must produce every document responsive to a request, and in patent search a single overlooked publication can invalidate a patent. Using software to find documents for this kind of review is known as technology assisted review, and has been studied in both legal and medical settings~\citep{cormack2014tar,kanoulas2017clef}. In all of these applications, missing a relevant document is costly, but so is returning too many, because everything returned must still be read.

\begin{figure*}[t]
\centering
\includegraphics[width=\linewidth]{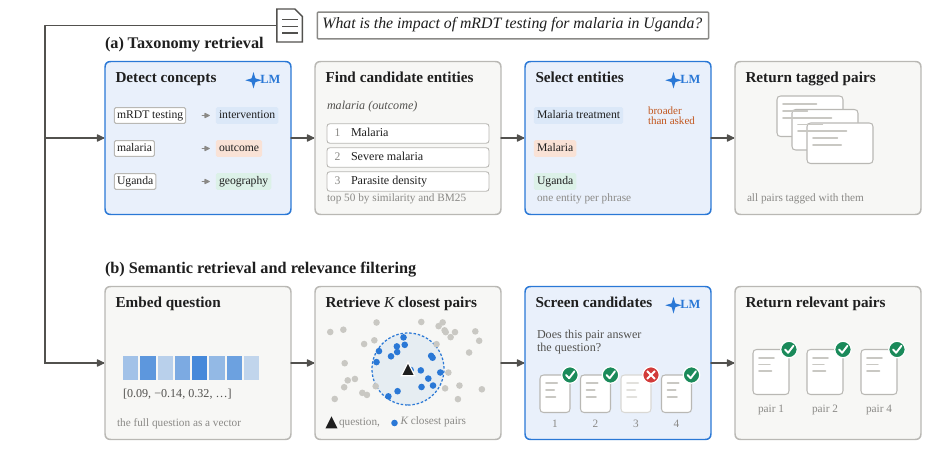}
\caption{The two retrieval designs, applied to the same question. (a) Taxonomy retrieval: the question is linked to entities in a catalog, and the pairs tagged with them are returned. (b) Semantic retrieval: the pairs closest to the question in embedding space are retrieved, then screened for relevance by a language model. Shaded steps marked LM call a language model.}
\label{fig:architectures}
\end{figure*}

These applications have traditionally relied on curated taxonomies. Each document is tagged with labels from a controlled vocabulary, each query is mapped to the same labels, and the documents that carry them are returned. Systematic review guidance still recommends this, combined with free text search~\citep{cochrane2025search}, for good reasons: labels normalize terminology, make each search inspectable, and group results for comparison and pooling. Until recently, taxonomies were also the only reliable option, since embedding retrievers~\citep{karpukhin2020dpr} performed poorly on specialized collections~\citep{thakur2021beir}.

The weakness of this approach is that the same labels serve both to find documents and to organize them. Labels compress every document and every query into a few categories, which makes them good for organizing but limits them for finding. A question narrower than its label returns everything under that label, so many of the results do not answer it. Worse, a concept missing from the vocabulary cannot be matched at all, and no amount of additional searching will find it. The share of relevant documents a taxonomy can find is therefore capped by the coverage of its vocabulary.

The alternative separates the two jobs. Documents and queries are embedded in the same vector space, the documents closest to the query are retrieved, and a language model then screens each of these candidates against the full question. This follows the multistage retrieval pattern, in which a fast model selects candidates and a slower, more accurate one examines them~\citep{wang2011cascade,nogueira2019bert}, here with a language model as the second stage~\citep{sun2023rankgpt}. Since the application needs a set of relevant documents and their order does not matter, the language model keeps or drops each candidate instead of reordering them. Taxonomy labels can still be applied afterward, to organize what was found.

We compare the two designs in ImpactAI,\footnote{\url{https://impactai.worldbank.org}} an evidence synthesis application built on 7,354 randomized controlled trials in development economics. We measure how much of the evidence included in 29 systematic reviews each design finds, how well the screening model agrees with human annotators, and how long each design takes to retrieve evidence for a query. The share of included studies found rises from 45\% to 85\%, and the screening model agrees with annotators as well as they agree with each other. In exchange, the median time to retrieve evidence for a query rises from 6 to 77 seconds. The number of candidates screened per query therefore becomes the main design choice, trading completeness against waiting time.

\section{Retrieval Designs for High Recall}\label{sec:designs}

This section describes the evidence ImpactAI retrieves, then the two retrieval designs, shown side by side in Figure~\ref{fig:architectures}.

\subsection{Evidence and Task}
ImpactAI is built on 7,354 studies across sectors including health, education, agriculture, public finance, and social protection, each reporting a randomized controlled trial (RCT). A study usually tests one or more interventions on several outcomes, so ImpactAI splits each study into intervention--outcome pairs, each holding the study's results for one intervention on one outcome (Table~\ref{tab:pairs}). The corpus contains 74,826 pairs, and the same structured evidence underlies Query2Effect, a benchmark for predicting effects from natural language questions~\citep{martinelli2026predicting}.

\begin{table}[t]
\centering\small
\begin{tabular}{@{}l>{\raggedright\arraybackslash}p{0.35\columnwidth}>{\raggedright\arraybackslash}p{0.43\columnwidth}@{}}
\toprule
Pair & Intervention & Outcome \\
\midrule
76717 & Malaria rapid diagnostic tests & Societal cost per 1,000 fever episodes \\
76718 & Malaria rapid diagnostic tests & Children with malaria treated appropriately \\
\bottomrule
\end{tabular}
\caption{Two pairs from one study, \emph{The impact of mRDT malaria tests in Uganda}. Each pair holds the study's results for that intervention and outcome. The descriptions indexed with each pair are omitted.}
\label{tab:pairs}
\end{table}

Pairs are the unit of retrieval. Given a question, the task is to return every pair that addresses it, and each returned pair carries its results, along with the population, context, and design of its study. Retrieving pairs instead of whole studies matters because a study often reports several results, of which only some answer a given question. Each pair also stays linked to its study, so the share of relevant evidence found can be measured in studies as well as in pairs.

\subsection{Taxonomy Retrieval}
\label{sec:taxonomy}
The first design, shown in Figure~\ref{fig:architectures}a, uses one taxonomy both to find and to organize evidence. Offline, every pair is tagged with entities, the concepts of a development economics catalog. At query time, a language model identifies the phrases in the question that name a concept and assigns each a type, such as intervention, outcome, population group, or geography. For each phrase, the system then retrieves up to 50 candidate entities of that type, combining semantic similarity with BM25 keyword matching~\citep{robertson2009bm25}, and a language model selects the entity that best fits the phrase in the context of the full question. Finally, the pairs tagged with the selected entities are returned.

This design has the benefits described in Section~\ref{sec:introduction}: results arrive grouped by shared categories, users can see which concepts their question was linked to, and a lookup returns every pair tagged with a given label. It also has the limits described there. In Figure~\ref{fig:architectures}a, for example, a question about rapid diagnostic tests for malaria is linked to the broader concept of malaria treatment, and returns all the evidence tagged with it. In addition, because the pipeline chains phrase detection, typing, entity selection, and tagging, an error at any step loses evidence. Finally, for a fixed catalog the result is a fixed set, so no amount of additional searching can add to it.

\subsection{Semantic Retrieval and Relevance Filtering}\label{sec:neural}

\begin{table*}[t]
\centering\small
\begin{tabular}{@{}lr>{\raggedright\arraybackslash}p{0.33\textwidth}lr@{}}
\toprule
Dataset & Questions & Items labeled & Labeled by & Relevant per question \\
\midrule
\multicolumn{5}{@{}l}{\textit{Retrieval quality}} \\
User questions & 374 & 1,000 closest pairs from Gemini and from Voyage, and taxonomy results & Language model & 150 pairs \\
Synthetic questions & 1,000 & 1,000 closest pairs from Gemini and from Voyage, and taxonomy results & Language model & 141 pairs \\
\midrule
\multicolumn{5}{@{}l}{\textit{Screening agreement}} \\
User questions & 100 & 30 closest pairs & One annotator & 24.0 pairs \\
Synthetic questions & 100 & 30 closest pairs & One annotator & 23.8 pairs \\
User questions & 10 & 30 closest pairs & Two annotators & --- \\
\midrule
\multicolumn{5}{@{}l}{\textit{Full pipeline}} \\
Systematic reviews & 348 & Included studies in the corpus & Review authors & 4.07 studies \\
User feedback & 45 & Pairs returned by semantic retrieval & Users & 17 pairs \\
\bottomrule
\end{tabular}
\caption{Evaluation datasets, grouped by the question each answers. Relevant counts for screening agreement include pairs labeled potentially relevant.}
\label{tab:datasets}
\end{table*}

The second design, shown in Figure~\ref{fig:architectures}b, separates finding from organizing. Each pair is represented by its text, the intervention and outcome names and descriptions together with the study context, which is embedded once, offline. At query time, the question is embedded into the same space, and the $K$ pairs closest to it form the candidate set, where $K$ is a fixed number. Retrieval thus draws on the text of the evidence itself, with no taxonomy in between.

Closeness in the embedding space, however, is only a proxy for relevance. Close candidates often share a topic with the question but differ in the intervention, outcome, population, or context the user asked about. A language model therefore screens each candidate: given the question and a pair's text, it decides whether the pair answers the question, and pairs that do not are discarded.

Screening is applied only to the $K$ closest pairs because it is far more expensive than retrieval. Comparing a question with every precomputed embedding is fast, whereas each screening decision requires a language model call at query time. Retrieval therefore narrows the corpus to a number of candidates that screening can afford, following the usual logic of a cascade: a cheap step over everything, then an expensive step over a few. Section~\ref{sec:latency} measures how this cost grows with $K$.

The two steps play distinct roles, which is why we evaluate them separately. Retrieval sets the ceiling on what can be found: screening cannot recover a pair that was never retrieved, and when more than $K$ pairs are relevant, some are necessarily missed. Screening then decides how much of the retrieved evidence is kept, and how much irrelevant material remains. Taxonomy labels can be applied to the resulting set afterward, to group and compare what was found.

\section{Experimental Setup}\label{sec:setup}

Following Section~\ref{sec:designs}, we evaluate the semantic design one step at a time, then as a whole. We first measure retrieval quality, which bounds what screening can keep, then how well screening agrees with human judgment, then how much evidence the full pipeline recovers, and finally how long it takes. Table~\ref{tab:datasets} summarizes the datasets used at each step. Two kinds of questions recur throughout: user questions, submitted to ImpactAI during its beta, and synthetic questions, each generated from one study in the corpus.

\subsection{Models}
We compare four embedding models for retrieval: Gemini embeddings~\citep{lee2025gemini}, Voyage~\citep{voyage}, E5-large-v2~\citep{wang2024textembeddingsweaklysupervisedcontrastive}, and Granite~\citep{granite}. Screening uses Gemini 2.5 Flash~\citep{comanici2025gemini25pushingfrontier}, which taxonomy retrieval also uses for phrase detection and entity selection. The deployed taxonomy service timed in Section~\ref{sec:latency} uses Gemini 3 Flash Preview instead.

\subsection{Retrieval Quality}
Retrieval is evaluated on 374 user questions and 1,000 synthetic questions. Labeling every pair in the corpus for each question is infeasible, so relevance is judged on a pool of candidates. For each question, the pool combines the 1,000 closest pairs under Gemini embeddings, the 1,000 closest under Voyage, and all pairs returned by taxonomy retrieval, and a separate language model, Gemini 3 Pro, labels each pooled pair for relevance. Scores are therefore relative to this pool. Because E5 and Granite did not contribute to it, relevant pairs that only they retrieve go unlabeled, which may understate their scores~\citep{zobel1998reliable}.

For each embedding model, we report three scores among the $K$ closest pairs, for $K$ between 10 and 1,000: recall, the share of relevant pairs retrieved; precision, the share of retrieved pairs that are relevant; and F1, which combines the two. Taxonomy retrieval returns the same set at any $K$, so its scores do not vary with depth.

\begin{table*}[t]
\centering
\resizebox{\linewidth}{!}{%
\begin{tabular}{@{}lcccccccccccccccccc@{}}
\toprule
 & \multicolumn{3}{c}{$K=10$} & \multicolumn{3}{c}{$K=100$} & \multicolumn{3}{c}{$K=150$} & \multicolumn{3}{c}{$K=300$} & \multicolumn{3}{c}{$K=500$} & \multicolumn{3}{c}{$K=1{,}000$} \\
\cmidrule(lr){2-4}\cmidrule(lr){5-7}\cmidrule(lr){8-10}\cmidrule(lr){11-13}\cmidrule(lr){14-16}\cmidrule(lr){17-19}
Model & R & P & F1 & R & P & F1 & R & P & F1 & R & P & F1 & R & P & F1 & R & P & F1 \\
\midrule
\multicolumn{19}{@{}l}{\textit{User questions}} \\
Taxonomy retrieval & \textbf{.15} & .36 & \textbf{.21} & .15 & .36 & .21 & .15 & .36 & .21 & .15 & \textbf{.36} & .21 & .15 & \textbf{.36} & .21 & .15 & \textbf{.36} & .21 \\
Gemini embeddings & .09 & \textbf{.65} & .16 & \textbf{.47} & \textbf{.52} & \textbf{.49} & \textbf{.62} & \textbf{.43} & \textbf{.51} & \textbf{.84} & .33 & \textbf{.47} & \textbf{.90} & .27 & \textbf{.41} & \textbf{.93} & .13 & \textbf{.24} \\
Voyage & .05 & .36 & .09 & .27 & .26 & .27 & .33 & .23 & .27 & .46 & .17 & .25 & .55 & .13 & .21 & .67 & .09 & .15 \\
E5-large-v2 & .05 & .33 & .08 & .19 & .20 & .20 & .23 & .17 & .20 & .32 & .13 & .18 & .39 & .09 & .15 & .50 & .06 & .11 \\
Granite & .05 & .40 & .09 & .26 & .26 & .26 & .32 & .23 & .26 & .42 & .17 & .24 & .50 & .12 & .20 & .62 & .08 & .14 \\
\midrule
\multicolumn{19}{@{}l}{\textit{Synthetic questions}} \\
Taxonomy retrieval & \textbf{.24} & .61 & \textbf{.35} & .24 & \textbf{.61} & .35 & .24 & \textbf{.61} & .35 & .24 & \textbf{.61} & .35 & .24 & \textbf{.61} & .35 & .24 & \textbf{.61} & \textbf{.35} \\
Gemini embeddings & .14 & \textbf{.81} & .24 & \textbf{.53} & .54 & \textbf{.53} & \textbf{.65} & .48 & \textbf{.55} & \textbf{.84} & .35 & \textbf{.49} & \textbf{.92} & .24 & \textbf{.38} & \textbf{.96} & .13 & .23 \\
Voyage & .09 & .46 & .15 & .34 & .29 & .31 & .41 & .25 & .31 & .53 & .19 & .28 & .63 & .14 & .24 & .76 & .09 & .17 \\
E5-large-v2 & .09 & .41 & .15 & .35 & .30 & .33 & .33 & .19 & .24 & .42 & .11 & .18 & .49 & .11 & .17 & .60 & .07 & .13 \\
Granite & .09 & .44 & .14 & .30 & .25 & .28 & .36 & .22 & .27 & .46 & .16 & .24 & .55 & .12 & .20 & .67 & .08 & .14 \\
\bottomrule
\end{tabular}}
\caption{Retrieval quality: recall (R), precision (P), and F1 among the $K$ closest pairs, judged on the pooled labels. Taxonomy retrieval returns a fixed set, so its scores do not change with $K$. Best score per column in bold.}
\label{tab:retrieval}
\end{table*}

\subsection{Screening Agreement}
Screening is evaluated on 100 user questions and 100 synthetic questions. For each question, an annotator labeled the 30 pairs closest to it under Gemini embeddings as relevant, potentially relevant, or not relevant. Most of these candidates are on topic, with 24.0 and 23.8 of the 30 labeled relevant or potentially relevant on average, so the test focuses on the fine distinctions screening has to make. Gemini 2.5 Flash screens the same pairs, and we compare its decisions with the human labels, counting only \emph{relevant} as positive.

To put this agreement in context, ten additional user questions were labeled independently by two annotators, and we measure their agreement on the 276 of 300 pairs that received both labels. Annotators were assigned per question, so the two annotators in Table~\ref{tab:screening} are not the same two people throughout. In both comparisons, we report Cohen's $\kappa$, which measures agreement beyond what chance would produce, along with Pearson correlation and F1 scores.

\subsection{Full Pipeline}
The full pipeline is evaluated against systematic reviews, which serve only as a reference and play no part in the system. The authors of a systematic review search the literature exhaustively and decide, study by study, which ones answer the review's question. Many of the studies they include are also in ImpactAI, so asking the system a review's question reproduces the search a review team would run, with the authors' decisions as the expert answer.

From 29 reviews, we derived 348 questions: the questions the reviews address, and narrower versions of them. The relevant studies for each question are the studies the review included that are also in the ImpactAI corpus. Reviews were not required to have all their studies in the corpus; included studies outside it are simply left out of the reference. For the narrower questions, a language model helped decide which of these studies apply. Each question has 4.07 relevant studies on average.

Returned pairs are mapped to their studies, and we report the share of relevant studies found, averaged over questions (macro recall), for $K$ of 100, 500, 1,000, and 2,500. Results at smaller $K$ are computed from the 2,500 closest pairs. We also report a combination of the two designs, which adds the taxonomy results, unscreened, to the screened semantic results. A review's list contains only the studies its authors included, so a returned study outside the list is not counted as an error. These lists therefore measure recall but not precision.

Precision is measured instead through user feedback. On 45 questions, users rated the pairs returned by semantic retrieval in the deployed application with an upvote or a downvote, and we report the share rated relevant. % CHECK: was screening on? Also, 17 relevant pairs per question (about 765 in total) conflicts with the 348 assessed pairs in Section 4.

\subsection{Latency}\label{sec:latency-setup}
We time each module on 30 questions from the systematic review benchmark, run one at a time: taxonomy retrieval, semantic retrieval alone, and semantic retrieval with screening, at the same four values of $K$. Taxonomy retrieval and $K=2{,}500$ are timed on all 30 questions, while times at smaller $K$ are estimates calibrated on three of them. In the timed configuration, screening groups candidates from the same study, up to 15 per prompt, with a separate decision for each pair. This configuration is more recent than the one used in the recall and agreement experiments. We report the mean, median, and 95th percentile.

\section{Results}\label{sec:results}

\begin{table}[t]
\centering\small
\setlength{\tabcolsep}{4pt}
\begin{tabular}{@{}lccccc@{}}
\toprule
 & & & \multicolumn{3}{c}{F1} \\
\cmidrule(l){4-6}
Comparison & $\kappa$ & Pearson & All & Rel. & Not rel. \\
\midrule
\multicolumn{6}{@{}l}{\textit{User questions (100)}} \\
Model vs.\ annotator & .50 & .50 & .75 & .76 & .74 \\
\multicolumn{6}{@{}l}{\textit{Synthetic questions (100)}} \\
Model vs.\ annotator & .50 & .51 & .75 & .75 & .75 \\
\multicolumn{6}{@{}l}{\textit{User questions (10)}} \\
Annotator vs.\ annotator & .43 & .47 & .71 & .67 & .74 \\
\bottomrule
\end{tabular}
\caption{Screening agreement. The model is Gemini 2.5 Flash; only pairs labeled relevant count as positive. F1 is reported over all pairs and for each class.}
\label{tab:screening}
\end{table}

\subsection{Retrieval Quality}
Among the embedding models, Gemini embeddings retrieve the most complete candidate sets at every depth (Table~\ref{tab:retrieval}). On user questions, Gemini recall reaches 0.47 at $K=100$ and 0.93 at $K=1{,}000$, against 0.67 at $K=1{,}000$ for Voyage, the next best model. Synthetic questions give the same ranking, with Gemini recall reaching 0.96 at $K=1{,}000$. We therefore use Gemini embeddings in the remaining experiments.

Taxonomy retrieval, by contrast, finds only a small share of the relevant pairs, and searching deeper cannot raise it: its fixed set recovers 15\% of relevant pairs on user questions and 24\% on synthetic ones. On user questions, Gemini at $K=100$ already does better on both recall (0.47 against 0.15) and precision (0.52 against 0.36). On synthetic questions, the taxonomy keeps higher precision (0.61 against 0.54 at $K=100$), but its recall stays far lower.

Going deeper, however, brings in irrelevant pairs much faster than relevant ones. On user questions, Gemini precision falls from 0.52 at $K=100$ to 0.13 at $K=1{,}000$, so most of a deep candidate set is irrelevant. This is the gap screening has to close.
% E5 Synthetic K=100: author corrected R=.35 and P=.30; source F1=.33 retained.
% F1 is author-reported, not recomputed from rounded P/R. Its aggregation and
% recall decrease from .35 at K=100 to .33 at K=150 remain to be reconciled.

\subsection{Screening Agreement}
Screening agrees with human annotators about as well as annotators agree with each other (Table~\ref{tab:screening}). Gemini 2.5 Flash reaches a Cohen's $\kappa$ of 0.50 on both user and synthetic questions, against 0.43 between two annotators, and an overall F1 of 0.75, against 0.71. In absolute terms, agreement is moderate on both sides, which shows how hard these fine distinctions are to judge, even for people. The comparison between annotators rests on only ten questions, and clearer annotation guidelines could reduce this disagreement.

\begin{figure}[t]
\centering
\includegraphics[width=\linewidth]{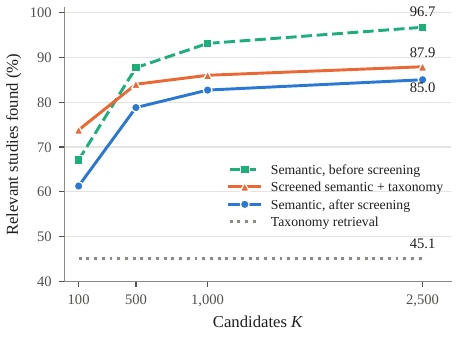}
\caption{Share of relevant studies found on the systematic review benchmark (macro recall), by number of candidates $K$. Screened semantic + taxonomy adds the taxonomy results, unscreened, to the screened semantic results.}
\label{fig:recall}
\end{figure}

\subsection{Full Pipeline}
Semantic retrieval with screening finds nearly twice as many of the studies included in systematic reviews as taxonomy retrieval: 85.0\% at $K=2{,}500$, against 45.1\% (Figure~\ref{fig:recall}). The gain comes from retrieval: before screening, the candidates already contain 96.7\% of the relevant studies. Screening then discards some relevant studies along with irrelevant ones, lowering recall by 11.7 points.

Recall grows with the number of candidates, but with diminishing returns. After screening, it rises from 61.3\% at $K=100$ to 82.7\% at $K=1{,}000$, and only to 85.0\% at $K=2{,}500$, although the last step more than doubles the candidates to screen.

The two designs also find partly different studies. Adding the taxonomy results, unscreened, to the screened semantic results raises recall by 12.6 points at $K=100$, to 73.8\%, but by only 2.9 points at $K=2{,}500$, to 87.9\%. The taxonomy is therefore most useful as a complement when the candidate budget is small.

Finally, the review lists measure recall but not precision. In user feedback on 45 questions, users rated 94.7\% of the returned pairs as relevant, so the pairs that reach users are mostly on target.
% CHECK: number of rated pairs (348 in the source, or 17 per question) and whether screening was on.

\begin{table}[t]
\centering\small
\setlength{\tabcolsep}{4pt}
\begin{tabular}{@{}lrrrr@{}}
\toprule
Module & $K$ & Mean & Median & 95th pct. \\
\midrule
Taxonomy retrieval & --- & 6.87 & 6.00 & 10.05 \\
\midrule
Semantic retrieval & 100 & 1.11 & 1.11 & 1.12 \\
 & 500 & 1.18 & 1.17 & 1.22 \\
 & 1,000 & 1.26 & 1.25 & 1.35 \\
 & 2,500 & 1.52 & 1.48 & 1.73 \\
\midrule
Semantic retrieval & 100 & 7.01 & 6.94 & 7.53 \\
and screening & 500 & 18.98 & 18.66 & 21.59 \\
 & 1,000 & 33.95 & 33.30 & 39.17 \\
 & 2,500 & 78.86 & 77.22 & 91.91 \\
\bottomrule
\end{tabular}
\caption{Time per question, in seconds, for each module. Times at $K$ below 2,500 are estimates calibrated on three questions; the others are measured on 30.}
\label{tab:latency}
\end{table}

\subsection{Latency}\label{sec:latency}
Screening accounts for almost all of the time, and that time grows with the number of candidates (Table~\ref{tab:latency}). Semantic retrieval alone takes between 1.1 and 1.5 seconds per question at the median, whatever $K$. With screening, the median rises from 6.9 seconds at $K=100$ to 33.3 seconds at $K=1{,}000$ and 77.2 seconds at $K=2{,}500$, against 6.0 seconds for taxonomy retrieval. At this rate, screening all 74,826 pairs would take roughly 40 minutes per question, which is why retrieval must narrow the set first.

Read together with recall, these times show what depth costs. At $K=100$, semantic retrieval with screening takes about as long as taxonomy retrieval (6.9 against 6.0 seconds) and finds more of the relevant studies (61.3\% against 45.1\%). Going from $K=1{,}000$ to $K=2{,}500$, by contrast, adds 2.3 points of recall and 44 seconds per question. Because the two services use different models (Section~\ref{sec:setup}), these times compare deployed modules rather than the designs themselves.

\section{Discussion}\label{sec:discussion}

\paragraph{Use meaning to find evidence and labels to organize it.}
Semantic retrieval with screening finds nearly twice as many of the studies included in systematic reviews as taxonomy retrieval. Because it matches the full question against the text of the evidence, it can find concepts that are missing from the catalog or narrower than its labels, which is the likely source of the gain. Labels remain useful once the evidence is found: they normalize terminology, make results inspectable, and group them for comparison, which pooling requires. They also remain the right tool when a user wants everything under a known category, and at small candidate budgets they add studies that semantic retrieval misses. The two designs are therefore best used together, each for the job it does well.

\paragraph{Depth is the main design choice.}
With semantic retrieval in place, the remaining choice is how many candidates to screen. Each additional candidate costs time while recall gains shrink, so a comprehensive search, such as one feeding a meta-analysis, may justify a long wait, whereas an exploratory search can use a smaller budget. Two directions could lower this cost: setting the depth for each question, since broad questions have more relevant pairs than narrow ones, and screening with smaller, faster models.

\paragraph{Maintenance and scale.}
The two designs also differ in upkeep. A taxonomy must be extended and the evidence retagged as the collection grows, while semantic retrieval only needs new embeddings, along with monitoring of the screening model, whose decisions are harder to audit as a whole than explicit labels. As the collection grows, broad questions will also have more relevant pairs, so a fixed budget will miss a larger share of them, and depth, with its cost, will need to grow too.

\section{Conclusion}\label{sec:conclusion}
Applications such as systematic reviews, legal discovery, and patent search need every relevant document, and a taxonomy caps how many it can find. In ImpactAI, embedding retrieval followed by language model screening raised the share of systematic review studies found from 45\% to 85\%, with screening that agrees with people as well as people agree with each other, at the cost of a longer wait that the number of candidates controls. Our evidence comes from one collection, but the mechanism is general: wherever labels are used to find documents, separating finding from organizing is likely to raise recall in the same way.

\section*{Limitations}
Our results come from one collection of RCTs in development economics, and both the taxonomy labels and the text of each pair depend on the quality of the information extracted from each study. The systematic review benchmark measures how many of a review's included studies are found within this collection, not coverage of the literature as a whole, and its narrower questions rely on study assignments made with the help of a language model. The retrieval benchmarks are labeled by a Gemini model over a pool built partly from Gemini embeddings, which may favor them~\citep{zobel1998reliable}, and the human samples are small, particularly the ten questions labeled by two annotators. The latency comparison covers deployed modules with different configurations: the timed taxonomy service uses Gemini 3 Flash Preview, and the timed screening configuration is more recent than the one used to measure recall and agreement. Throughput under concurrent load, monetary cost, and the effect of more complete evidence on pooled estimates remain to be measured. Finally, the exact versions of Voyage and Granite used were not recorded.

\section*{Acknowledgments}
We are deeply grateful to the Development Impact AI Lab's ground truth experts, Nihaa Sajid, Clémence Gall, Rida Ali Khan, Saqib Hussain, Aarushi Aggarwal, and Ejigayehu Diriba. Their relevance judgments are the human reference against which screening is measured, and this evaluation would not have been possible without them. We also thank Google.org for its generous support of this work.

\bibliography{refs}

\end{document}